\documentclass[accepted]{uai2024} 

\usepackage{amsmath,amsfonts,bm}

\def\eqref#1{equation~\ref{#1}}

\def\1{\bm{1}}

\DeclareMathAlphabet{\mathsfit}{\encodingdefault}{\sfdefault}{m}{sl}
\SetMathAlphabet{\mathsfit}{bold}{\encodingdefault}{\sfdefault}{bx}{n}

\usepackage[american]{babel}

\usepackage{hyperref}
\usepackage{url}
\usepackage{graphicx}
\usepackage{caption}

\usepackage{algorithm}
\usepackage{algorithmic}
\usepackage{setspace}
\usepackage{listings}

\usepackage{upquote}  
\usepackage{xcolor,colortbl}

\usepackage{natbib} 

\usepackage{booktabs} 
\usepackage{siunitx} 

\lstdefinestyle{overleaf}{
    backgroundcolor=\color[rgb]{0.95,0.95,0.92},   
    commentstyle=\color[rgb]{0,0.6,0},
    keywordstyle=\color{magenta},
    numberstyle=\tiny\color[rgb]{0.5,0.5,0.5},
    stringstyle=\color[rgb]{0.58,0,0.82},
    basicstyle=\ttfamily\footnotesize,
    breakatwhitespace=false,         
    breaklines=true,                 
    captionpos=b,                    
    keepspaces=true,                 
    numbers=left,                    
    numbersep=5pt,                  
    showspaces=false,                
    showstringspaces=false,
    showtabs=false,                  
    tabsize=2
}
\lstdefinestyle{simple}{
  backgroundcolor=\color{white},
  basicstyle=\fontsize{7.5pt}{7.5pt}\ttfamily\selectfont,
  keywordstyle=\fontsize{7.5pt}{7.5pt}\ttfamily\selectfont,
}
\title{AcceleratedLiNGAM:\\Learning Causal DAGs at the speed of GPUs}

\author[1]{Victor Akinwande}
\author[1,2]{J. Zico Kolter}
\affil[1]{%
    Carnegie Mellon University
}
\affil[2]{%
    Bosch Center for AI
}

\begin{document}
\maketitle

\begin{abstract}
  Existing causal discovery methods based on combinatorial optimization or search are slow, prohibiting their application on large-scale datasets.  In response, more recent methods attempt to address this limitation by formulating causal discovery as structure learning with continuous optimization but such approaches thus far provide no statistical guarantees. In this paper, we show that by efficiently parallelizing existing causal discovery methods, we can in fact scale them to thousands of dimensions, making them practical for substantially larger-scale problems. In particular, we parallelize the LiNGAM method, which is quadratic in the number of variables, obtaining up to a 32-fold speed-up on benchmark datasets when compared with existing sequential implementations. Specifically, we focus on the causal ordering sub-procedure in DirectLiNGAM and implement GPU kernels to accelerate it. This allows us to apply DirectLiNGAM to causal inference on large-scale gene expression data with genetic interventions yielding competitive results compared with specialized continuous optimization methods, and VarLiNGAM for causal discovery on U.S. stock data.
  
\end{abstract}

\section{Introduction}
Causal discovery or inference aims to learn causal interactions in a data-driven way \citep{pearl2000models, imbens2015causal}. However, most causal discovery methods need to establish the causal relationships between every pair of variables in the data, which results in a complexity that is at least quadratic in the number of variables \citep{glymour2019review}. For instance, DirectLiNGAM \citet{shimizu2011directlingam, hyvarinen2013pairwise} works by recursively performing regression and conditional independence tests between pairs of variables to establish the causal ordering and then iteratively regressing each variable on other variables higher in the established causal order. Although DirectLiNGAM allows for full determination of the causal structure under minimal assumptions of linearity, non-Gaussian errors, acyclicity, and the absence of hidden confounders, it has a complexity of O($d^3$) where $d$ is the number of variables.

To address this scalability limitation, a recent trend in the literature has been to formulate heuristics combined with continuous optimization or deep learning methods to learn the underlying structure of the data \citep{lee2019scaling, Lopez2022largescale, montagna2023scalable}. However, care must be taken when ascribing any causal interpretation to such methods. While these methods enhance computational efficiency, and allow us to apply causal discovery to large datasets, they rely on restrictive assumptions, are sensitive to hyper-parameters and more importantly, their identifiability guarantees are often not established. In-fact, as we show in the sections that follow, NOTEARS \citet{zheng2018dags} - a widely used method  cannot recover the underlying structure in a simple causal DAG. As such, there is a compelling need to explore other avenues to achieve scalability of causal discovery methods.

In this paper, we describe an implementation of the LiNGAM analysis, which is accelerated by parallelization on consumer GPUs. We focus on the DirectLiNGAM and VarLiNGAM methods, but the ideas presented are easily applicable to other LiNGAM variants. Parallelization leads to a \textbf{32-fold} speed-up over the sequential version, enabling us to apply these methods to large-scale datasets.
We apply DirectLiNGAM to gene expression data with genetic interventions ($d \approx 1000$) and VarLiNGAM to U.S. stock data ($d = 487$).
We observe that there is still a need to explore improved ways to speed up LiNGAM on GPUs, such as better I/O awareness and increasing the proportion of the algorithm that is parallelized. Our implementation is open-sourced on GitHub\footnote{\url{https://github.com/Viktour19/culingam}}.

\section{Background}
Causal Directed Acyclic Graphs (DAGs) are a fundamental structure in causal inference and graphical models.
A causal DAG consists of nodes, each representing a random variable. These variables can be anything of interest, such as different attributes, events, or states. The edges in a causal DAG are acyclic, directed and represent causal relationships between the variables. An edge from variable $X_i$ to $X_j$ implies $X_i$ has a direct causal influence on $X_j$. Similarly, the absence of a direct edge from $X_i$ to $X_j$ indicates that, according to the model, $X_i$ does not have a direct causal effect on $X_j$ given the other variables and edges in the model. Causal DAGs encode the assumptions about the conditional independence of the variables meaning the joint probability distribution over all possible values of the variables can be factorized into a product of conditional distributions given by:
$$P(X_1, X_2, ..., X_n) = \prod_{i=1}^{n} P(X_i | \text{Parents}(X_i))$$

In this section, we provide a brief overview of Functional Causal Models based methods for learning causal DAGs, we discuss the GPU execution model and how it is amenable to such methods, we present the basic LiNGAM analysis and discuss recent methods in the literature that seek to learn DAGs using heuristics and continuous optimization.

\subsection{Causal discovery based on Functional Causal Models (FCM)}
 FCMs represent a causal effect $Y$ as a function of direct causes $X$, noise $\epsilon$, and parameter sets $\theta$, assuming that the transformation from causes to effects is invertible e.g Eqn.~\ref{eq:fcm}. For instance, in the linear, non-Gaussian acyclic model (LiNGAM) \citet{shimizu2006linear}, the causal direction can be determined when at most one of the noise term or cause is Gaussian. The post-nonlinear (PNL) causal model extends this by considering nonlinear effects such as sensor or measurement errors, and is identifiable in most cases \citet{zhang2012identifiability}. FCMs provide a flexible framework for causal discovery by capturing the causal asymmetry in the data-generating process and accommodates various data distributions. When the exact functional form of the data-generating process is unknown or in cases with discrete variables and small cardinality, properties of the causal process may be obscured and precise identification of causal directions remains challenging.

\begin{equation}
Y = f(X, \epsilon; \theta_1)
\label{eq:fcm}
\end{equation}


\subsection{Standard LiNGAM implementation}
LiNGAM \citet{shimizu2006linear} is characterized by properties that enable identification of causal relationships between variables. The observed variables $x_i$, $i \in\{i, \ldots m\}$ in the dataset can be arranged in a causal order denoted by $k(i)$. This order implies that no later variable in the sequence causes any earlier variable, forming a recursive structure that can be represented as a DAG. This property also ensures there are no cyclic relationships between the variables. In addition, the value of $x_i$ is determined linearly by the values of the variables before it in the causal order i.e. $x_i = \sum_{k(j) < k(i)} \theta_{ij} x_j + \epsilon_i + c_i$ where $\theta_{ij}$ represent the strength of the causal effect of $x_j$ on $x_i$, $\epsilon_i$ is the noise term and $c_i$ is an optional constant term. The final property is that the noise terms $\epsilon_i$ are mutually independent, continuous random variables with non-Gaussian distributions.

\begin{figure}[ht!]
\begin{minipage}{0.24\textwidth}
  \includegraphics[width=\linewidth]{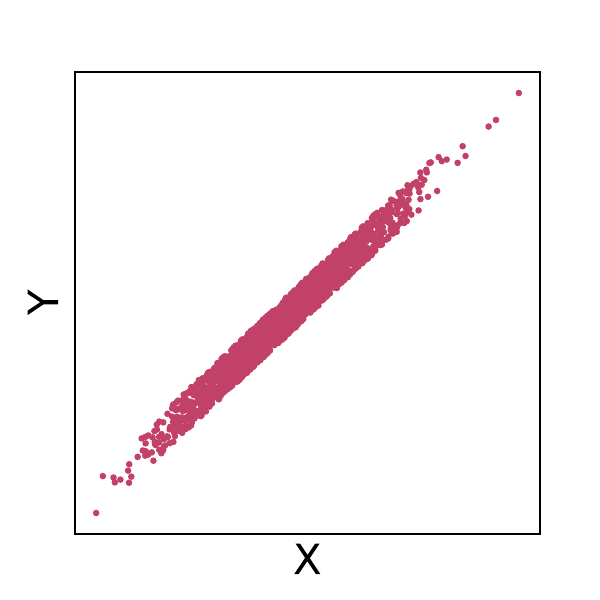}
\end{minipage}%
\begin{minipage}{0.24\textwidth}
  \includegraphics[width=\linewidth]{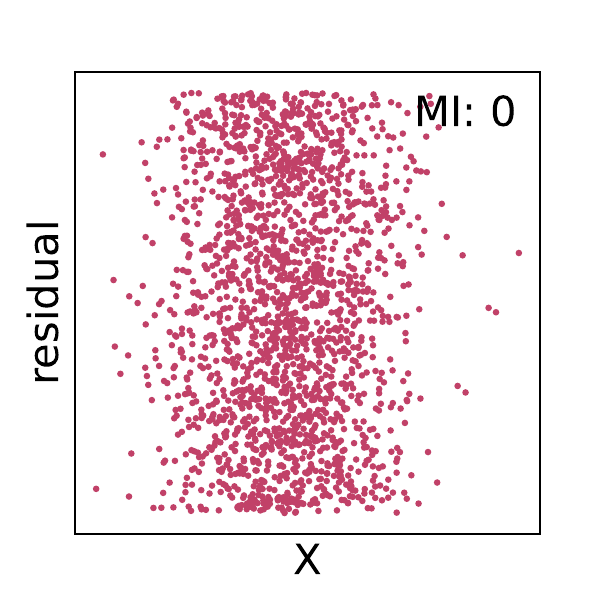}
\end{minipage}
\begin{minipage}{0.24\textwidth}
  \includegraphics[width=\linewidth]{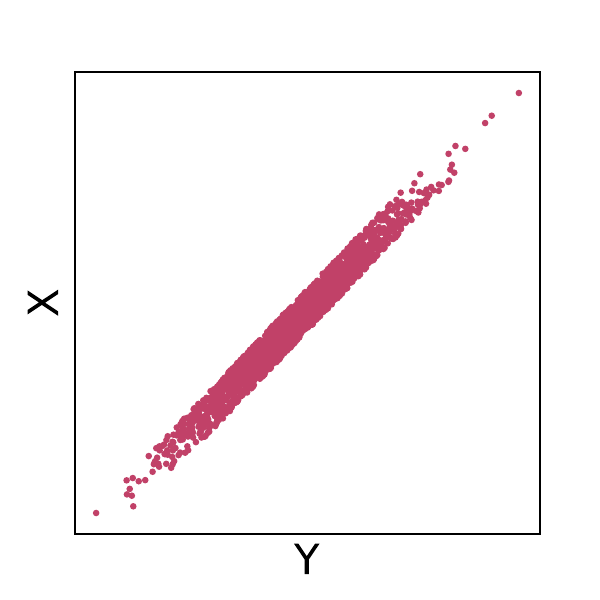}
\end{minipage}%
\begin{minipage}{0.24\textwidth}
  \includegraphics[width=\linewidth]{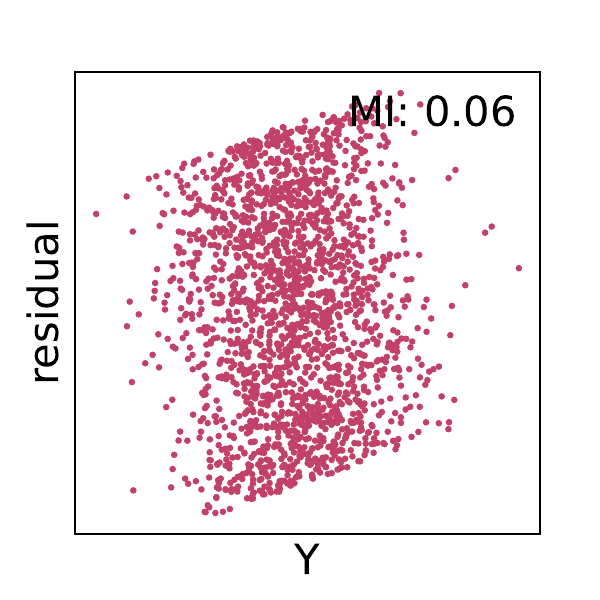}
\end{minipage}
\caption{Illustration of the causal asymmetry principle underpinning LiNGAM. Given data generated according a LiNGAM functional causal model as in Eqn.~\ref{eq:fcm}, the regression residual can only be independent of the independent variable in the correct causal direction (top figure). This holds for any distribution of the noise except Gaussian. Independence is measured using the Mutual Information (MI).}
\label{fig:assymetry}
\end{figure}

Constrained by these assumptions on the data-generating process, the independence and non-Gaussianity of $\epsilon$ provides valuable information needed to establish the causal direction say given a pair of variables. This is because we can easily identify independence between the cause and the error terms. As such, any causal model that is inconsistent with this identification is discarded. We illustrate this principle in Figure \ref{fig:assymetry}. Given data generated according a LiNGAM functional causal model as in Eqn.~\ref{eq:fcm}, the regression residual can only be independent of the independent variable in the correct causal direction.

In practice, datasets consist of more than two variables, so we need a general method for estimating the causal order for datasets of arbitrary dimensions. In \citet{shimizu2011directlingam}, DirectLiNGAM is presented as a method to achieve this. DirectLiNGAM begins by identifying a variable not caused by any other variable (termed exogenous) in the data through its independence from the residuals of multiple pairwise regressions. The effects of this variable are then removed from the other variables using least squares regression. \citet{shimizu2011directlingam} showed that a LiNGAM structure also holds for the residuals of the regression, meaning that the residuals themselves can be treated as variables in a LiNGAM model. In addition, the causal ordering of the residuals corresponds to the causal ordering of the original observed variables. DirectLiNGAM therefore iteratively applies the same independence and removal steps to the residuals, finding the ``exogenous'' residual at each iteration. By repeatedly identifying and removing the effects of exogenous variables (or residuals), the model estimates the causal order of the original variables. We show the implementation to identify the variable at position $k(i)$ in Algorithm \ref{alg:causalorder}.


\begin{algorithm}[t]
\caption{\text{Causal ordering pseudo-implementation.}}\label{alg:causalorder}
\newcommand{\hlbox}[1]{%
  \fboxsep=1.2pt\hspace*{-\fboxsep}\colorbox{black!10}{\detokenize{#1}}%
}
\lstset{style=simple}
\begin{lstlisting}[language=Python]
# X      : dataset [m, dim]
# U      : indices of variables [dim]
# _residual: computes the regression residual
# _diff_mutual_info: computes the difference of the mutual information

def search_causal_order(X, U):
k_list = np.zeros(len(U))
for i in U:
    k = 0
    for j in U:
        if i != j:
            xi = X[:, i]
            xj = X[:, j]
            xi_std = (xi - np.mean(xi)) / np.std(xi)
            xj_std = (xj - np.mean(xj)) / np.std(xj)

            ri_j = _residual(xi_std, xj_std)
            rj_i = _residual(xj_std, xi_std)

            mi_diff = _diff_mutual_info(xi_std, xj_std, ri_j, rj_i)
            k += np.minimum(0, mi_diff) ** 2
            
    k_list[i] = -k
return U[np.argmax(k_list)]
\end{lstlisting}
\end{algorithm}

\subsection{GPU execution model}
\begin{figure}[ht!]
\begin{minipage}{0.25\textwidth}
  \includegraphics[width=\linewidth]{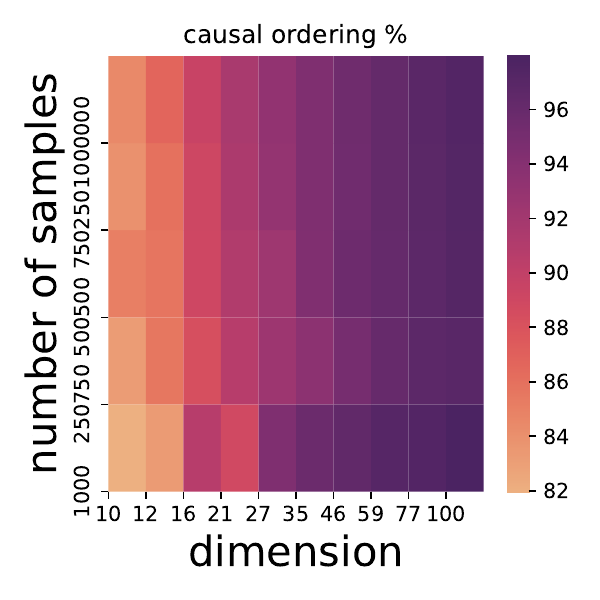}
\end{minipage}%
\begin{minipage}{0.25\textwidth}
  \includegraphics[width=\linewidth]{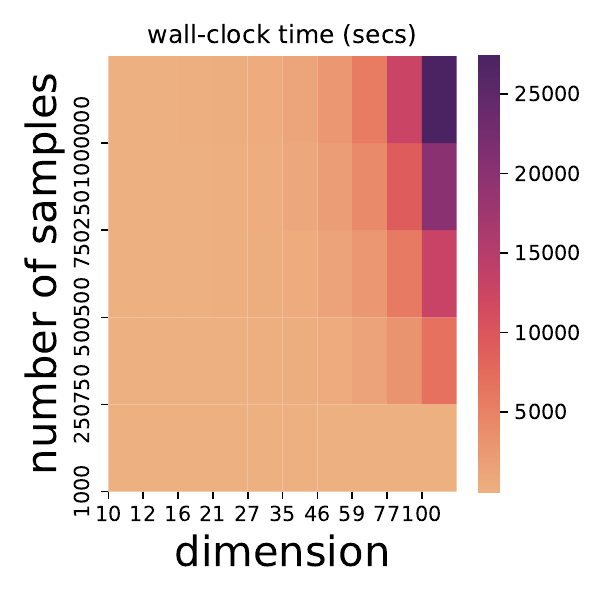}
\end{minipage}
\begin{minipage}{0.25\textwidth}
  \includegraphics[width=\linewidth]{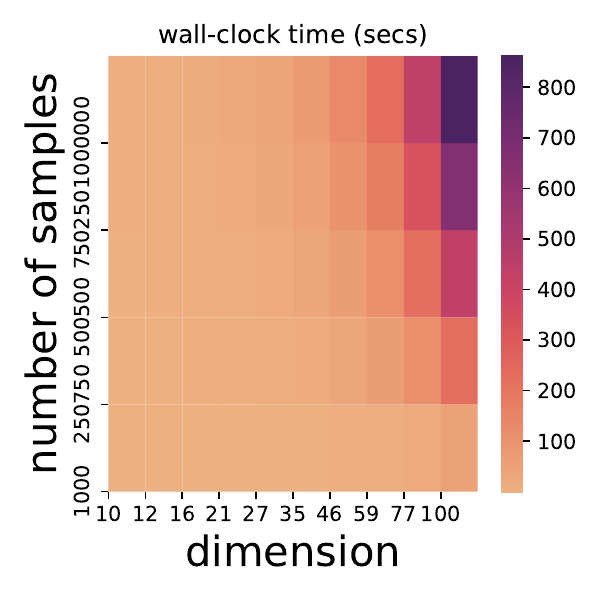}
\end{minipage}%
\begin{minipage}{0.25\textwidth}
  \includegraphics[width=\linewidth]{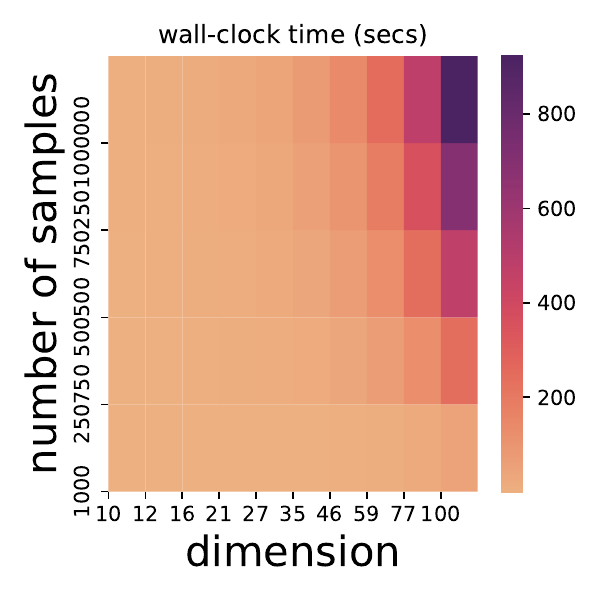}
\end{minipage}
\caption{Benchmark of CPU (sequential) implementation of DirectLiNGAM. Given data with specified number of samples and dimensions, the causal ordering sub-procedure accounts for up to \num{96}\% of overall runtime \textbf{(top-left)}. It takes \num{7} hours on a CPU to process a dataset of \num{1} million samples with \num{100} variables \textbf{(top-right)}. Benchmark of GPU (parallel) implementation of DirectLiNGAM \textbf{(bottom-left)} and VarLiNGAM \textbf{(bottom-right)}. Given data with specified number of samples and dimensions, the parallel implementation achieves up to \num{32} times speed-up when compared to the sequential implementation. The benchmark is obtained using an NVIDIA RTX 6000 Ada with \num{18176} cores.}
\label{fig:props}
\end{figure}

The causal ordering procedure (Algorithm \ref{alg:causalorder}) is the main computational bottleneck of the DirectLiNGAM algorithm, accounting for up to \num{96}\% of the overall wall-clock time (see Figure \ref{fig:props}). The pseudo-implementation of the procedure also makes the limitation of DirectLiNGAM, in practice, glaring. The algorithm needs to compute statistical measures in an inner loop for each pair of variables in the data. Such computational structure results in a complexity quadratic in the number of variables making it difficult to apply on large-scale datasets. DirectLiNGAM takes \num{7} hours to process a dataset with \num{1} million observations and \num{100} variables on a high-performance AMD EPYC server CPU (see Figure \ref{fig:props}).

On a second look, we observe that each variables pair computation is independent of the others. This independence is a key characteristic that makes an algorithm suitable for parallel processing. In-fact, based on this, successful attempts to parallelize the algorithm on a super-computer have been made \citep{matsuda2022accelerating}. However, GPUs present a more accessible approach and have not be explored. GPUs can perform the same operation on multiple data-points simultaneously (vectorization), and also handle high arithmetic intensity computations. Operations involving accumulations can also be efficiently parallelized on GPUs by using techniques like parallel reduction, where the work is divided among multiple threads and then combined to get the final result. This implies vectorized computations like the mean and standard deviation can be done much faster and computation of residuals and mutual information differences for different pairs can be done in parallel batches. Finally, recent work \cite{shahbazinia2023paralingam} has demonstrated that LiNGAM is amenable to GPU acceleration, although the authors propose using a heuristic to prune the search procedure. This effectively modifies the algorithm, and the implementation is not available for us to benchmark.

\subsection{Continuous optimization based structure learning}
Learning DAGs with GPUs is certainly not a new idea. Continuous optimization methods for learning DAGs are amenable to acceleration on GPUs using packages like \textit{PyTorch} or \textit{Tensorflow}. Algorithms such as NOTEARS \citet{zheng2018dags} and GOLEM \citet{ng2020role} simultaneously optimize over the DAGs structure and its parameters by defining a differentiable acyclicity constraint and enabling end-to-end optimization of a score function over graph adjacency matrices. NOTEARS minimizes the MSE between the observations and the model predictions:
$$\text{MSE}_X(\theta) = \frac{1}{m} \|X - X\theta\|_F^2$$ where $\|\cdot\|_F$ denotes the Frobenius norm. NOTEARS includes a constraint defined with a trace exponential function that equals zero if and only if $\theta$ represents an acyclic graph, and a penalty term $\lambda\|\theta\|_1$ where $\lambda\|.\|_1$ is defined element-wise and $\lambda$ is a hyper-parameter.

GOLEM performs Maximum Likelihood Estimation (MLE) under the assumption of Gaussian noise terms in the data, and includes both soft acyclicity and sparsity constraints.

Unfortunately, both methods make restrictive assumptions about the data-generating process such as equal noise variance across observations, or that the marginal variance of each variable being strictly larger than its ancestors in the DAG (termed varsortability) \citep{reisach2021beware, ng2023structure}. The non-convex nature of the optimization problem in GOLEM often necessitates careful initialization and sophisticated optimization strategies to ensure convergence to a meaningful solution. Moreover, neither NOTEARS nor GOLEM provides identifiability guarantees; they may not perform reliably on simulated datasets where the true underlying structure is known (see Section \ref{sec:impl}), and other such methods may not converge \citep{Lopez2022largescale}. Furthermore, the effectiveness of these algorithms is highly sensitive to the choice of hyper-parameters, such as the sparsity threshold or the specific loss function employed. Selecting these hyper-parameters is non-trivial and may require extensive cross-validation or domain expertise, potentially limiting the practicality of these methods.
\section{AcceleratedLiNGAM: Analysis, and Extensions}
In this section, we discuss the considerations that make acceleration of Algorithm \ref{alg:causalorder} efficient on GPUs. First, note that the outer loop over $i$ and the inner loop over $j$ are independent and thus can be parallelized. The dependency on the results of \text{\_residual} and \text{\_diff\_mutual\_info} for each pair $(i, j)$ however, necessitates synchronization and memory management. Therefore, we parallelize over $i$, each block (group of threads) handles a different $i$ value, computing $\text{k\_list}[i]$ and within each $i$, we parallelize over $j$. We also ensure that within the inner loop, operations are ordered using GPU abstractions.

This parallelization scheme requires up to $dim * (dim - 1)$ cores. Profiling the sequential implementation (Figure \ref{fig:props}), Amdahl's law suggests a theoretical speed-up of \num{25}. In the limit of infinite processors, $speedup = 1/(1 - p)$, where $p$ is the parallelizable portion of the algorithm (\num{0.96} in our case) but this does not account for the increase in workload size with the number of processors \citep{gustafson1988reevaluating}. Finally, we do not need synchronization for updating $\text{k\_list}[i]$ since the update order does not matter.

\subsection{Efficient DirectLiNGAM implementation}\label{sec:impl}
We implemented AcceleratedLiNGAM and evaluate on an NVIDIA RTX 6000 Ada. The implementation optimizes for memory use by performing parallel reductions in shared memory. Shared memory refers to a fast on-chip memory space shared among the threads of a block in the GPU. The results show a 32-fold speed-up of DirectLiNGAM when compared with the sequential implementation (see Figure \ref{fig:props}).

\begin{figure}[ht!]
\centering
\begin{minipage}{0.40\textwidth}
  \includegraphics[width=\linewidth]{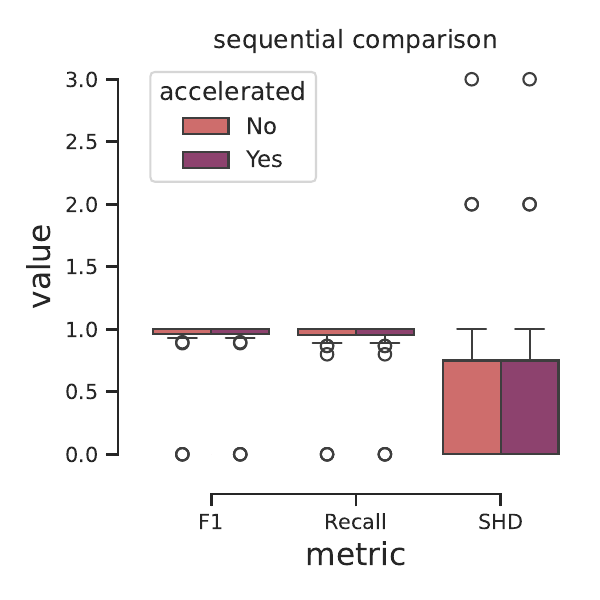}
\end{minipage}
\centering
\begin{minipage}{0.23\textwidth}
  \includegraphics[width=\linewidth]{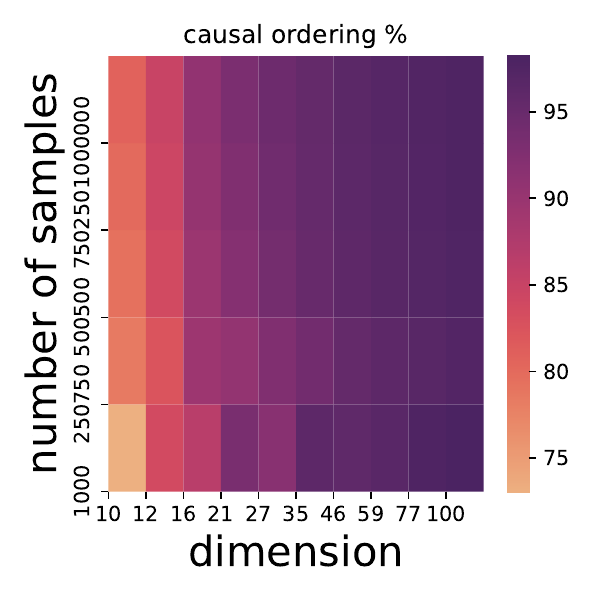}
\end{minipage}%
\begin{minipage}{0.23\textwidth}
  \includegraphics[width=\linewidth]{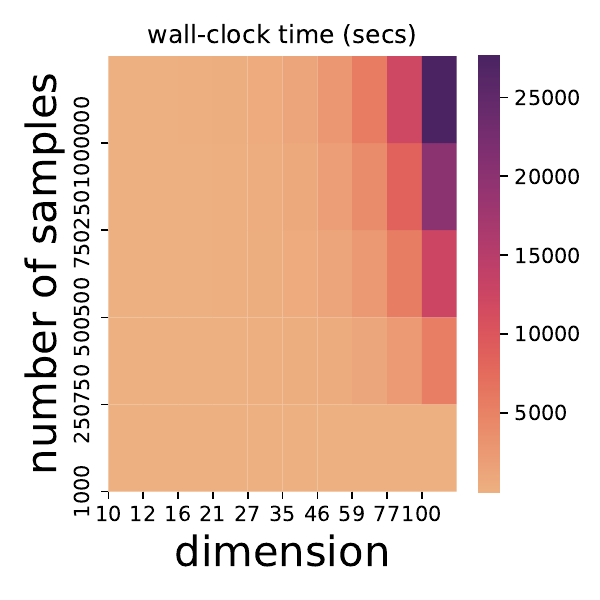}
\end{minipage}
\caption{Comparison of parallel and sequential implementation of DirectLiNGAM. We simulate data according to a linear FCM with \num{10000} samples, and \num{10} variables. Both implementations produce the exact same result \textbf{(top)}. Benchmark of CPU (sequential) implementation of VarLiNGAM. Given data with specified number of samples and dimensions, the causal ordering sub-procedure of DirectLiNGAM accounts for up to \num{96}\% of overall runtime \textbf{(bottom-left)}. It takes \num{7} hours on a CPU to process a dataset of \num{1} million samples with \num{100} variables \textbf{(bottom-right)}.}
\label{fig:comparison}
\end{figure}

To validate that there are no logical errors in our parallel implementation, we compare the results of applying the sequential implementation with those of the parallel implementation on simulated data. We report the F1 score, recall, and structural hamming distance (SHD) over \num{50} simulations (different random seed) in Figure \ref{fig:comparison}.

Let $G = (V, E)$ be a DAG where $V$ is a set of vertices representing variables and $E$ is a set of directed edges representing causal relationships between variables. Vertices are connected such that each vertex $v_i$ at level $l$ may have parents from the set of nodes at level $l -1$:
$\forall v_i \in V, \exists l_i \in \{0, 1\} \text{ such that } (v_j, v_i) \in E \Rightarrow l_j = l_i - 1$. Given $G$, we generate data such that the strength of the causal effect $\theta \sim \mathcal{N}(0, 1)$, and the noise terms $\epsilon_i \sim \text{Uniform}(0, 1), \quad \text{for } i = 1, 2, ..., m$. Comparison of the sequential and parallel implementation of DirectLiNGAM on this simulated data show that they produce the exact same result, and recover the true causal graph accurately (see Figure \ref{fig:comparison}).

We evaluate NOTEARS on similarly simulated data selecting the best performance across a grid: $\{0.001, 0.005, 0.01, 0.05, 0.1\}$ of $\lambda$ values. We obtain an F1 score of \num{0.79} $\pm$ \num{0.2}, Recall of \num{0.69} $\pm$ \num{0.2} and SHD of \num{2.52} $\pm$ \num{1.67}. This shows that even on data where the causal influences are simple, NOTEARS does not perform well.

\subsection{Extension: Efficient VarLiNGAM implementation}
The basic LiNGAM analysis can be extended to auto-regressive modeling. In \citet{hyvarinen2010estimation}, VarLiNGAM is introduced. The key idea is to combine FCMs and vector autoregressive (VAR) models to capture both instantaneous and lagged influences among multiple time series. The model is expressed as: $x(t) = \sum_{\tau=0}^{k} \theta_{\tau} x(t-\tau) + \epsilon(t)$
where $x(t)$ is the observed time series at time $t$, $\theta_{\tau}$ is the matrix of causal effects with time lag $\tau$ and $\epsilon(t)$ is the noise term (innovation). Similar to all LiNGAM analyses, the model assumes the noise terms are mutually independent, non-Gaussian, and the matrix $\theta_{\tau}$ corresponding to instantaneous effects forms an acyclic graph.

On a high-level, VarLiNGAM works by first estimating coefficients of the VAR model using standard auto-regressive modeling techniques, and then transforms the estimated VAR coefficients based on the causal adjacency matrix estimated using DirectLiNGAM thereby accounting for the direct causal relationships while isolating the indirect effects captured by the VAR coefficients. We refer the reader to \citet{hyvarinen2010estimation} for more details. The key thing to note here is the same algorithm in Algorithm \ref{alg:causalorder} dominates the runtime of VarLiNGAM (See Figure \ref{fig:comparison}) and so we obtain a similar speed-up of \num{30} with the GPU implementation.

\subsection{Low level CUDA implementation details}
All the benchmark results are run on an NVIDIA A6000 Ada GPU with \num{48} GB of memory, \num{18176} cores and \num{91.1} TFLOPS single-precision performance. The register file size is 64K 32-bit registers per Streaming Multiprocessor (SM), the maximum number of registers per thread is \num{255}, the maximum number of thread blocks per SM is \num{16}, the shared memory capacity per SM is \num{100} KB, and the maximum shared memory per thread block is \num{99} KB.

A key optimization is how the kernel uses shared memory for intermediate reduction results within each block. This pattern is sensitive to the number of threads because it assumes that the shared memory array is fully populated, and after storing in shared memory, we perform a reduction within the block. We set shared memory per thread block to \num{96} KB. We implement warp tiling (for optimal latency, and efficient synchronization) and notice a \num{20}\% speed-up but due to the non-associative nature of floating-point operations, naive implementation of parallel reduction may lead to rounding errors which we observe in the residual computation. As such, we leave this optimization for future work.

Finally, optimizing GPU implementations of machine learning methods for I/O awareness has achieved significant success recently, especially when combined with the use of cores optimized for fast matrix multiplication (Tensor cores) \cite{dao2022flashattention, fu2023flashfftconv}. A profile of our GPU kernels reveals that the two most time-consuming operations are \textit{poll} and \textit{pthread\_cond\_timedwait}, both accounting for about \num{50}\% of the execution time. These operations are associated with waiting for I/O operations or synchronization primitives, indicating substantial room for optimization to enhance I/O awareness, and efficient synchronization. The remainder of the DirectLiNGAM and VarLiNGAM implementations, which are not parallelized, involve several regression analysis. Although, we utilize heavily optimized libraries like \textit{numpy} and \textit{scikit-learn} for these regressions, there remains potential for speed-up through parallelism with Tensor cores.

\section{Experiments}
\subsection{AcceleratedLiNGAM to gene expression data with genetic interventions}
We experiment with the causal learning of gene regulatory networks from gene expression data, with genetic interventions, as detailed in \citet{friedman2000using,pe2001inferring,Lopez2022largescale}. This approach is enabled by a single-cell RNA sequencing technique known as Perturb-Seq, as described by \citet{dixit2016perturb} . Perturb-Seq allows for targeted genetic interventions and the subsequent measurement of their effects on the complete gene expression profiles in hundreds of thousands of individual cells using single-cell RNA-seq.

Our experimental setup mirror that of \citet{Lopez2022largescale}, where the Perturb-CITE-seq dataset \citet{frangieh2021multimodal} contains expression profiles from \num{218331} melanoma (cancer) cells, after interventions targeting each of \num{249} genes. For each gene in the genome, measurement from a single-cell combines the identity of the intervention (the target gene) along with a count vector, that is the expression level of a particular gene. The dataset includes patient-derived melanoma cells with the same genetic interventions but exposed to three conditions: co-culture with T cells derived from the patient's tumor (\num{73114} cells) (these can recognize and kill melanoma cells), interferon (IFN)-$\gamma$ treatment (\num{87590} cells) and control (\num{57627} cells). We retain cells from \num{20}\% of the interventions as a test set. The dimensions (samples, dim) of the train set for co-culture, IFN, and control datasets are (\num{65164}, \num{964}), (\num{75443}, \num{964}), and (\num{50539}, \num{961}) respectively.

We applied AcceleratedLiNGAM to each of the three datasets. Since LiNGAM analysis does not involve causal inference, after obtaining the weighted adjacency matrix, we apply standard Variational Inference (VI) methods to obtain both the interventional NLL (I-NLL) and the mean absolute error (I-MAE) across held-out interventions. Specifically, we use Stein VI \citet{liu2016stein} implemented in the \textit{Pyro} package where we defined a model such that variables without direct causal influence on another variable are leaf nodes and otherwise are latent nodes with priors $\sim \mathcal{N}(0, 1)$. We generate \num{200} posterior samples, and optimize for \num{5000} iterations. DirectLiNGAM itself is not sensitive to the random seed. 

On a high-level, Stein VI involves approximating the target distribution \(p(x)\) by iteratively updating a set of particles. This process minimizes the KL divergence between the approximating distribution \(q(x)\) and the target \(p(x)\), which is achieved through the application of smooth transforms. The key insight is to apply a perturbation to the identity map, \(T(x) = x + \epsilon \phi(x)\), where \(\phi(x)\) is a smooth function that characterizes the perturbation direction, and \(\epsilon\) is a small scalar representing the perturbation magnitude. This approach ensures \(T\) is an injective (one-to-one) map, maintaining the full rank of the Jacobian of \(T\), and thereby guaranteeing the invertibility required for the change of variables formula to hold.

\begin{table}[h]
\centering
\caption{Comparison of DirectLiNGAM with VI and DCD-FG method on the Perturb-CITE-seq datasets. We obtain the I-NLL (nll) and I-MAE (mae) on all three datasets. Lower values are better.}
\label{tab:my_label}
\begin{tabular}{@{}lSSSSSS@{}}
\toprule
& \multicolumn{2}{c}{\textbf{Co-culture}} & \multicolumn{2}{c}{\textbf{IFN}} & \multicolumn{2}{c}{\textbf{Control}} \\
\cmidrule(r){2-3} \cmidrule(lr){4-5} \cmidrule(l){6-7}
& {nll} & {mae} & {nll} & {mae} & {nll} & {mae} \\
\midrule
DirectLiNGAM & 1.5 & 0.7 & 1.5 & 0.9 & 3 & 1.6 \\
DCD-FG ($\approx$) & 1.1 & 0.7 & 1.2 & 0.7 & 1.1 & 0.7 \\
\bottomrule
\label{tbl:infmetrics}
\end{tabular}
\end{table}

We compare the results from DirectLiNGAM, combined with VI, with those obtained using DCD-FG introduced in \citet{Lopez2022largescale}. DCD-FG works by combining a parameterized distribution over factor directed graphs with a hybrid likelihood model, optimized with an acyclicity constraint and was applied to the Perturb-CITE-seq datasets. We observe the I-MAE to be lower or about the same on the co-culture dataset (one leaf variable), and slightly higher on the IFN and control datasets (one and two leaf variables respectively) (see Table~\ref{tbl:infmetrics}. We note DCD-FG is a continuous optimization based structure learning method and prone to many of the issues we have previously discussed but more pertinently, the results presented in \citet{Lopez2022largescale} do have quite a bit of variance. We also observe the I-NLL of DirectLiNGAM to be slightly higher on all datasets. 
While this may seem like worse performance, it is interesting to note how the control dataset (i.e., no interventions) has the same I-NLL as the other two datasets with DCD-FG. However, with DirectLiNGAM, the I-NLL of the co-culture and IFN datasets are similar, but the control I-NLL is much higher. Since there is no ground truth data, it is difficult to determine if DCD-FG is overfitting but it seems likely.

\subsection{AcceleratedLiNGAM to stock data with auto-regression modelling}

We also experiment with learning the causal relationships among U.S. stock indices. We obtain hourly adjusted closing price data for the stocks in the S\&P 500 from Yahoo Finance from January 2022 to December 2023. We preprocess the data lightly, filling missing values using time-based linear interpolation, removing indices with any remaining missing values, and transforming the original series to a new series that is stationary with first differencing before applying VarLiNGAM. This process leaves us with data from \num{487} indices. Finally, we consider a lag of \num{1} in the VAR model and use the estimator provided by the \textit{statsmodels} package.

\begin{figure}[ht!]
\begin{minipage}{0.48\textwidth}
  \includegraphics[width=\linewidth]{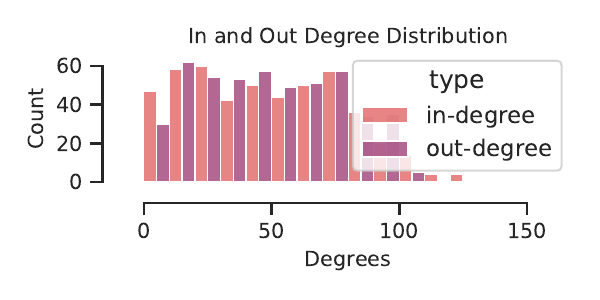}
\end{minipage}
\caption{In and out degree distribution of the adjacency matrix obtained using VarLiNGAM on S\&P 500 hourly data. We observe some level of symmetry between in-degree and out-degree, and the fairly uniform distribution across a range of degrees indicates that there are no prominent hubs that significantly stand out.}
\label{fig:degrees}
\end{figure}

In Figure~\ref{fig:degrees}, we show the in and out degrees distribution of the adjacency matrix we obtain ($\theta_0$). The distribution of in-degrees and out-degrees is similar, which could imply that the instantaneous causal relationships have some form of balance in terms of influences made and received by the nodes. The graph also has two leaf nodes corresponding to \textit{USB}: U.S. Bancorp and \textit{FITB}: Fifth Third Bancorp which is interesting because both companies are holding companies i.e. they do not engage directly in business operations themselves but rather hold the assets of and oversee the management of their subsidiaries. Consequently, the discovery that they are influenced by other indices without influencing any in return is particularly relevant.

We obtain the top 5 exerting casual influence nodes, and the top 5 receiving influence nodes by measuring the total causal effects. We show the indices that correspond to both categories in Table~\ref{tbl:infl}. The top exerting indices predominantly features companies involved in manufacturing, retail, and services with a direct consumer focus, such as homebuilding (NVR), auto parts retail (AZO), and food services (CMG). These companies are often more directly impacted by consumer spending habits and economic cycles, given their close ties to discretionary spending (e.g., travel, dining out, and home purchases). Whereas, the top receiving indices operate in industries subject to more stringent regulations or external factors, like CenterPoint Energy (CNP) in the utilities sector, or that play a crucial role in the information and packaging industries, such as News Corp (NWSA) and Amcor plc (AMCR), where changes in consumer behavior or regulatory environments can significantly impact their operations.


\begin{table}[h]
\centering
\caption{We show the indices that correspond to both the top 5 exerting casual influence nodes, and the top 5 receiving influence nodes from the discovered causal graph.}
\begin{tabular}{ll}
\toprule
\multicolumn{2}{c}{Top Exerting} \\
\midrule
NVR$_{\tau-1}$  & Homebuilding \\
AZO$_{\tau-1}$  & Auto Parts \\
CMG$_{\tau-1}$  & Restaurants \\
BKNG$_{\tau-1}$ & Online Travel \\
MTD$_{\tau-1}$  & Precision Instruments \\
\midrule
\multicolumn{2}{c}{Top Receiving} \\
\midrule
NWSA$_{\tau}$ & Diversified Media \\
CNP$_{\tau}$  & Utilities \\
FOXA$_{\tau}$ & Broadcasting \\
AMCR$_{\tau-1}$ \& AMCR$_{\tau}$ & Packaging \\
\bottomrule
\label{tbl:infl}
\end{tabular}
\end{table}


\section{Conclusion}

Machine learning practitioners often encounter use-cases where predictive performance alone does not suffice. Typical examples include in healthcare, genomics, or predictive justice applications where the understanding of cause-and-effect relationships is paramount. In healthcare, accurately predicting patient outcomes is crucial, but understanding the causal factors behind diseases can lead to more effective treatments and health policies \citep{raisanen2006causation, barros2022causal}. Similarly, in genomics, identifying the causal relationships between genetic markers and diseases is vital for developing targeted therapies and personalized medicine approaches \citep{burgess2018inferring, Lopez2022largescale}. In predictive justice applications, it is not enough to predict recidivism rates; understanding the causal factors can inform more effective rehabilitation programs and fairer justice policies \citep{khademi2020algorithmic, loeffler2022impact}.
In all these domains, the focus is not solely on prediction but also on uncovering the underlying causal mechanisms. By addressing the scalability limitations of causal discovery methods with statistical guarantees, we aim to enable the widespread application of causal inference in large-scale data analysis.

In this paper, we accelerate LiNGAM methods, achieving significant speed-up compared to sequential implementations. We do not change the algorithm; hence, the identifiability guarantees of AcceleratedLiNGAM remain the same as those of the original methods. We note that there is still room for further speed-up through the use of specialized CUDA techniques and by leveraging the GPU memory hierarchy. Also, the performance of GPUs continues to increase over time. Thus, we anticipate that future iterations of AcceleratedLiNGAM could see even greater improvements in computational efficiency. This work opens up the application of LiNGAM to a wider range of applications, and by open-sourcing our implementation, we aim to spur the development of these improvements and applications by the broader community.

\section{Acknowledgments}
We thank Daiyaan Arfeen for valuable discussions. Victor Akinwande was supported by funding from the Bosch Center for Artificial Intelligence.

\newpage

\bibliography{tmlr}



\end{document}